\documentclass[conference, 10pt]{IEEEtran}
\IEEEoverridecommandlockouts
\usepackage{cite}
\usepackage{amsmath,amssymb,amsfonts, amsthm}

\usepackage{authblk}

\usepackage{graphicx}

\usepackage{algorithm}
\usepackage{algpseudocode}

\usepackage{siunitx}
\usepackage{textcomp,mathcomSTEv4}

\theoremstyle{plain}

\usepackage[utf8]{inputenc}

\usepackage[dvipsnames]{xcolor}

\usepackage{caption}
\usepackage{subcaption}

\usepackage{tikz,pgfplots}
\usetikzlibrary{calc}
\usepgfplotslibrary{fillbetween}
\pgfplotsset{compat=1.10}

\usepackage{pgfplotstable}
\usetikzlibrary{positioning}
\usetikzlibrary{pgfplots.groupplots}
\usetikzlibrary{arrows}
\usetikzlibrary{matrix}
\usepackage{multirow} 
\usepackage{amsthm,enumitem}

\usepackage{xcolor}
\def\BibTeX{{\rm B\kern-.05em{\sc i\kern-.025em b}\kern-.08em
    T\kern-.1667em\lower.7ex\hbox{E}\kern-.125emX}}

\usepackage{hyperref}
\hypersetup{
    colorlinks=true,
    linkcolor=black,
    filecolor=magenta,      
    urlcolor=cyan,
    pdfpagemode=FullScreen,
    }

\newtheorem*{remark}{Remark}

\newcommand{\gennorm}{\mathsf{GenNorm}}
\newcommand{\coiii}{\mathsf{CO}_3}

\begin{document}

\title{
Convert, compress, correct:  Three steps toward  communication-efficient DNN training
}


\author{
Zhong-Jing Chen\textsuperscript{1}
}
\author{
Eduin E. Hernandez\textsuperscript{2}
}
\author{
Yu-Chih Huang\textsuperscript{3}
}
\author{
Stefano Rini\textsuperscript{4}
}
\affil{National Yang-Ming Chiao-Tung University (NYCU), Taiwan}
\affil{\textit {\{\textsuperscript{1}zhongjing.ee10, \textsuperscript{2}eduin.ee08, \textsuperscript{3}jerryhuang, \textsuperscript{4}stefano.rini\}@nycu.edu.tw}}

\maketitle

\begin{abstract}
In this paper,  we introduce a novel algorithm, $\coiii$, for communication-efficiency distributed Deep Neural Network (DNN) training.
%
%
$\coiii$ is a joint training/communication protocol which encompasses three processing steps for the network gradients:
(i) quantization through floating-point \underline{conversion}, (ii) lossless \underline{compression}, and (iii) error \underline{correction}.
These three components are crucial in the implementation of distributed DNN training over rate-constrained links.
The interplay of these three steps in processing the DNN gradients is carefully balanced to yield a robust and high-performance scheme. 
$\coiii$ is shown having better accuracy and improved stability, despite the reduced payload.
The performance of the proposed scheme is investigated through numerical evaluations over CIFAR-10. 
\end{abstract}


\section{Introduction}
The training of accurate and robust Deep Neural Networks (DNN) rely on the availability of large datasets.
As the complexity of modern-day DNN are ever-increasing, high performance is attainable only by training the network over a tremendous amount of data.
In this training regime, data centralization is no longer  feasible 
%
for two main reasons: On one hand the volume of data is too large to be communicated and stored centrally 
and, on the other hand, the centralization presents too many concerns from the privacy and security standpoint. 
%
%
For the reasons above,  distributed DNN training has received much attention in the recent literature, both  from a distributed
parallel, 
and networked 
computing  perspective.

In this paper, we consider the distributed DNN training in which a centralized model is trained over datasets present at remote users. 
For this scenario, we propose $\coiii$, a novel training/communication scheme where the gradients at the remote users are first (i) converted to low-resolution floating point representations, then (ii) compressed, losslessly, using the assumption that gradients can be well modelled as  generalized normal ($\gennorm$) i.i.d. samples, and then communicated to the PS through a finite-capacity link. At the next iteration, the remote users  (iii) correct the quantization error by adding a version of it to the current gradient before (i) is repeated in the next iteration.
%
%
%
%
We show that this approach provides excellent training performances at very low transmission rates between the remote user and the PS.

\smallskip 
\noindent
\underline{\bf Relevant Literature:}
Among various distributed optimization frameworks, 
federated learning (FL) has received particular attention in the recent literature \cite{Shalev-Shwartz2010FL_CE,Wang2018Spars_FL,Alistarh2018Spars_FL}.
FL consists of a central model which is trained   locally at the remote clients by applying Stochastic Gradient Descent (SGD) over a local dataset.
The local gradient are then communicated to the central Parameter Server (PS) for aggregation into a global model.
A natural constraint in distributed and decentralized optimization is with respect to transmission rates between nodes and its relationship to the overall accuracy \cite{saha2021decentralized,shlezinger2020communication}.
Accordingly, one is interested in devising rate-limited communication schemes that attain high accuracy at a low overall communication payload.
This can be attained through two steps  (i) dimensionality reduction and (ii)  quantization and compression. 
The dimensionality-reduction schemes put forth in the literature rely on various sparsification approaches
\cite{Shalev-Shwartz2010FL_CE,Alistarh2018Spars_FL}.
%
%
Following dimensionality reduction, the gradient can be digitized through quantization, either scalar-wise \cite{seide2014onebitSGD,Konecny2016Fl_CE,salehkalaibar2022lossy} or vector-wise \cite{gandikota2019vqsgd}.
From an implementation-oriented perspective, \cite{sun2019hybrid} studies the effect of gradient quantization when constrained to a \emph{sign-exponent-mantissa} representation.
%
%
%
%
After quantization, enabled by the statistical model obtained via extensive simulations that gradients in DNN training with SGD follows i.i.d. $\gennorm$, lossless compression can be applied to further reduce the communication rate toward the PS \cite{chen2021dnn}. 

When  gradients are compressed, it has been shown that error correction,  or error feedback, can greatly improve performance \cite{karimireddy2019error}. 
Error feedback for $1$-bit quantization was originally considered in \cite{seide20141}.
In \cite{stich2018sparsified}, error feedback is applied to gradient compression. 
\smallskip
\noindent
\underline{\bf Contributions:}
In this paper, we consider the problem of efficient gradient compression for rate-limited distributed DNN training. 
In particular, we expand our work of \cite{chen2021dnn} to include further mechanisms to improve the training accuracy.
The proposed scheme, which we term $\coiii$  is comprised of the three following gradient processing steps:  

\begin{enumerate}[ labelwidth=!, labelindent=-5pt,label=(\roman*)]
  \item  \underline{Floating Point (fp) {\bf Conversion}:} As a quantization mechanism, we consider fp conversion due its compatibility with gpu gradient processing.
  %
  \item \underline{Lossless Gradient {\bf Compression}:} 
  Quantized gradient undergo lossless compression. 
  \item \underline{Error {\bf  Correction}:} The quantization error is stored at one iteration and corrected in the next.
\end{enumerate}
We note that step (ii) requires a good statistical model for gradients with fp conversion and error correction. As in \cite{chen2021dnn}, our extensive simulations (only partially shown due to the lack of space) indicate that the i.i.d. $\gennorm$ model remains valid. With this assumption, we show that, by carefully designing the parameters of these three steps, $\coiii$  can attain high accuracy at a much reduced communication payload. 

\noindent
{\bf Notation.}
Lowercase boldface letters (e.g., $\zv$) are used for tensors, 
uppercase letters for random variables (e.g. $X$), and calligraphic uppercase for  sets (e.g. $\Acal$) .
We also adopt the short-hands  $[m:n] \triangleq \{m, \ldots, n\}$
and  $[n] \triangleq \{1, \ldots, n\}$. 
Both subscripts and superscripts letters (e.g. $g_t$ and $g^{(u)}$) indicate the iteration index or the user index for a tensor.
%
%
Finally, $\Fbb_2$ is the binary field.
The all zero vector is indicated as $\zv$.
\section{System Model}
\label{sec:System Model}

In many distributed training scenarios of practical relevance, the communication from the remote users and the PS is severely constrained in transmission rate.
For this reason, in the following,  we consider the approach of \cite{shlezinger2020communication,chen2021dnn} and consider the accuracy/payload trade-off of distributed optimization. 
%
%
In Sec. \ref{sec:DNN training} we specialized the general setting of \cite{shlezinger2020communication,chen2021dnn} to DNN training.
{
In order to obtain clearer insights on this trade
off, we wish to avoid the effects of the 
asynchronous training at multiple remote users. 
For this reason we consider the case in which training occurs simultaneously and synchronously at all remote users. 
}
%
%


\subsection{Distributed Optimization}
\label{sec:Distributed Optimization}

Consider the scenario with $U$ users, each possessing a local dataset
\ea{
\Dc^{(u)} = \left\{\lb \dv_{k}^{\lb u\rb},v_k^{\lb u\rb}\rb\right\}_{k\in\left[\left|\Dc^{(u)}\right|\right]},
}
where $\Dc^{(u)}$  includes
$\left|\Dc^{(u)}\right|$
pairs, each comprising of a data point 
$\dv_{k}^{\lb u\rb}$
and the label $v_{k}^{(u)}$
for $u \in [U]$. 
Users collaborate with the PS to minimize the loss function $\Lcal$ as evaluated across all the local datasets and over the choice of the model $\wv \in \Rbb^d$, that is 
\ea{
\Lcal(\wv) =  \f 1 {|\Dcal|} \sum_{u \in [U]} \sum_{k\in\left[\left|\Dcal^{(u)}\right|\right]} \Lcal(\wv_{t};\dv^{(u)}_{k},v^{(u)}_{k}).
\label{eq:loss}
}
%
For the \emph{loss function} in \eqref{eq:loss}, we assume that there exists a unique minimizer $\wv^*$. 
%
A common approach for numerically determining $\wv^*$ is through the iterative application of (synchronous) SGD.
In the SGD algorithm, the model parameter $\wv$ is updated at each iteration $t\in[T]$, 
by taking a step toward the negative direction of the gradient vector,   that is 
%
$\wv_{t+1}=\wv_{t}-\eta_t  \gv_t$
for $t \in [T]$, $\wv_{0}=\zerov_d$, and where  $\gv_t$ is the stochastic gradient of $\Lcal(\cdot)$, evaluated in $\wv_{t}$, 
with
$\Ebb\lsb \gv_t\rsb=  \nabla\Lcal\lb\wv_n\rb$.
 $\eta_t$ is  an iteration-dependent step size, called \emph{learning rate}.

In the FL setting, the SGD iterations  are distributed among $U$ users and is orchestrated by PS as follows:
(i) each user $u \in[U]$ receives the current model estimate, $\wv_t$ of the optimal model $\wv^*$ over the infinite capacity link from the PS.
The user $u \in [U]$ then (ii) accesses its  local dataset $\Dv^{(u)}$ and computes the (local) stochastic gradient $\gv_t^{(u)}$.
%
%
Finally (iii) each node communicates the gradient estimate $\gv_{t}^{(u)}$ to the PS which then computes the term $\gv_t$  
as
\ea{
\gv_t=\f{1}{U}\sum_{u \in[U]} \gv_t^{(u)},
\label{eq:aggregate}
}
and uses $\gv_t$ to update the model estimate.

\subsection{Rate-limited distributed training}
\label{sec:Rate-limited distributed training}

%
%
In the rate-limited distributed training scenario, 
communication between each user and the PS take place over a noiseless channel with finite capacity. 
On the other hand, the communication between the PS and the remote users takes place over a channel with infinite capacity.

A general three-step scheme to address the finite capacity constraint is described as follows.
%

First, the local gradient $\gv_{t}^{(u)}$ is  quantized via a quantizer $Q:\mathbb{R}\rightarrow \mathcal{X}$ to form the representative $\hat{\gv}_t^{(u)}=Q(\gv_{t}^{(u)})$, where $\mathcal{X}$ is the collection of representatives, i.e. quantization levels.
%
Following quantization, the quantized gradients are further compressed through the mapping $h:\mathcal{X}\rightarrow \mathbb{F}_2^{*}$ to form a codeword $\mathbf{b}_t^{(u)}=h(\hat{\gv}_t^{(u)})$. Here, we consider $h$ to be a variable-length coding scheme; hence, the range is $\mathbb{F}_2^{*}$, where $\mathbb{F}_2$ is the binary field.  
This compression step is lossless, that is the mapping $h$ is invertible: the role of this mapping is in  removing the statistical redundancy inherent in the local gradients, thus reducing the amount of bits to be transmitted to the PS. 
We shall utilize the following assumption for the design of the lossless compressor:
%
%
Let us assume that  the local gradient is distributed i.i.d. according to $\mathbb{P}_{\Gv_t}$ at each user, that is $\gv_t^{(u)} \sim \mathbb{P}_{G_t}$ is i.i.d for all $u \in [U]$. 
Using the assumption that a gradient distribution can be properly defined, we can then define the expectation of the compression performance.
More precisely, let $r_t^{(u)}$ be the length of $\mathbf{b}_t^{(u)}$. We define the expected length of $u\in[U]$ at $t\in[T]$ as
\ea{
R_t^{(u)} = \mathbb{E}_{Q,h} \left[r_t^{(u)}\right],
}
where the expectation is taken w.r.t. the gradient distribution at time $t$, $\Pbb_{\Gv_t}$.

At the PS, the gradient of user $u \in [U]$ at time $t$ is reconstructed $\ghv_t^{(u)}$  and the model is updated as
\ea{
\whv_t= \whv_{t-1}+  \f{\eta_t}{U}\sum_{u \in[U]} \ghv_t^{(u)},
\label{eq:aggregate hat}
}


Having introduced the problem formulation, we are now ready to define the {\em communication overhead} of a certain choice  of functions $(Q,h)$ as the  \underline{sum expected lengths} conveyed over the up-link channel  over the training, that is 
	\begin{equation}
	\label{eqn:Overhead}
	\Rsf = \sum_{t \in [T]} \sum_{u \in [U]} R_t^{(u)}.
	\end{equation}
	%
Using the definition in \eqref{eqn:Overhead}, we finally  come to the definition of the accuracy/overhead trade-off as 
%
\ea{
\Lsf_T(\Rsf) = \min_{Q,h} 
\Ebb\lsb \Lcal(\wv_T) \rsb,
\label{eq:Lsf}
}
where the expected value is over the stochasticity in the gradient evaluation and the distribution of the gradients.
In other words, $\Lsf_t(\Rsf)$ is the minimum loss in accuracy 
that one can attain  with respect to the unconstrained case when the total communication payload is $\Rsf$.
The  different between the loss in the unconstrained case, $\Lcal(\wv_t)$, and the constrained case, $\Lcal(\whv_t)$, is evaluated at iteration $T$, which is assumed to be the total number of iterations allowed for training.
Note  that the minimization is over the quantization and lossless compression operations.
%

\subsection{DNN training}
\label{sec:DNN training}
While we have so far considered a general distributed optimization problem, we shall focus our numerical evaluations in Sec. \ref{sec:Numerical evaluations} to the distributed DNN training problem. 
For distributed DNN training without error correction, we have argued and validated via extensive simulations in \cite{chen2021dnn} that the gradients distribute i.i.d. according to $\gennorm$ in each layer. Although we are still investigating theoretical validation of this hypothesis, we note that \cite{isik2021successive} adopts a similar assumption on the gradient distribution (Laplace instead of $\gennorm$).

\section{Proposed Approach: $\coiii$} 
\label{sec:Proposed Approach}

In the following we specialize the general scheme in Sec. \ref{sec:Rate-limited distributed training} to  our proposed approach,  $\coiii$.
More precisely, $\coiii$ considers the following gradient processing steps (i)  the quantization, $Q$, is chosen as the fp conversion, (ii) compression, $h$, is chosen as element-wise Huffman coding, and (iii) error correction is performed with a memory decay $\gamma$.

Next, let us detail each of the steps above in further detail. 

\smallskip
\noindent
{\bf (i) fp conversion:} the local gradient $\gv_t^{(u)}$ is converted into the fp representation with one bit for the sign, $\sf mant$ bits for the mantissa and $\sf exp$ bits for the exponent. Additionally, at each time $t$, we introduce a bias $b_t$ on the exponent so as to minimize the expected loss between the closest quantization representative and $\gv_t^{(u)}$, that is
\ea{
b_t = \argmin_b  \min_{\sf c_{\sf sgn}, c_{\sf mant}, c_{\sf exp}} \Ebb \lsb   c_{\sf sgn} \cdot c_{\sf mant}  \cdot 2^{ c_{\sf exp} +b} - G_t  \rsb,
\label{eq:exp bias}
}
where $\argmin$ is over $b_t \in \Rbb$, the $\min$ is over the sign, $c_{\sf sgn}$, and all possible values (decimal) of the mantissa and the exponent,  $c_{\sf mant}$ and $c_{\sf exp}$ respectively.
The expected value is over the gradient distribution $\Pbb_{\Gv_t}$.
Let us denote the fp quantization strategy as $Q_{\mathsf{fp}}(\cdot)$ in the following.

More principled approach to gradient quantization can be considered-- see \cite{salehkalaibar2022lossy}. 
Here we focus on fp conversion as it can be implemented with extreme computational efficiency.
%

\smallskip
\noindent
{\bf (ii) Huffman coding:} 
After fp conversion, the quantized gradient $\ghv^{(u)}_t$ is compressed using Huffman coding. 
%
As Huffman coding requires the distribution of data sources, we assume that the underlying distribution is $\gennorm$. 
Note that this assumption has been validated in \cite{chen2021dnn} for distributed DNN training without error correction. 
In Sec. \ref{sec:Numerical evaluations}, we will verify this assumption again through simulations and obtain corresponding parameters for distributed DNN with error feedback.
%
A different code is used at each DNN layer, but the same code is used across all users at a given layer.
Note that, as for the fp conversion, Huffman coding is chosen for $\coiii$  as it can be implemented  with  minimum requirements for both computation and memory.
This is in contrast with other universal compression algorithms, such as LZ74, which do not rely on any assumption on the source distribution.
A comparison in terms of gradient compression-ratio between these two algorithm can be observed in Sec. IV-D of \cite{chen2021dnn}.
In the following, we indicate Huffman lossless compression as $h_{\mathsf{Hf}}(\cdot)$.

\smallskip
\noindent
{\bf (iii) Error correction:}
Consider the error feedback strategy in Lines 8 to 10 of Algorithm \ref{alg:cap} where the quantization error is accumulated in the variable $\mv_t^{(u)}$.
As shown in the literature \cite{stich2018sparsified,stich2019error}, error feedback has been an effective tool in accelerating the convergence of models using compressed gradients, enabling a convergence rate which is comparable to their counterpart models that use the uncompressed gradients.
Apart from \cite{stich2018sparsified}, in Algorithm \ref{alg:cap}-Line 8, we introduce an additional parameter, $\gamma$, to discount the error accumulation.
We refer to this parameter as the \emph{memory decay coefficient}. 
In our numerical experimentations, a judicious choice of $\gamma$ will be proved crucial in tuning the performance. 

\begin{algorithm}
\caption{Proposed Algorithm: $\coiii$ }
\label{alg:cap}
\begin{algorithmic}[1]
\Require Local datasets $\{\Dcal_u \}_{u \in [U]}$, loss function $\Lcal(\cdot)$, initial model estimate $\whv_0$ 
\Require learning parameter $\eta$, memory decay parameter $\gamma$
\For{$u \in [U]$}
\State user $u$ sets memory to zero $\mv_0^{(u)}=\zerov$
\EndFor
\For{$t \in [T]$}
\State PS sends $\whv_t$ to all remote users
\For{$u \in [U]$}
\State user $u$ evaluates the local gradient $\gv^{(u)}_t$
\State user $u$  fp-converts $\gv^{(u)}$: $\ghv^{(u)}_t=Q_{\mathsf{fp}}(\gv^{(u)}_t+ \gamma \mv_{t-1}^{(u)} )$
\State user $u$ compresses  $\ghv^{(u)}_t$:  $\bv_t^{(u)}=h_{\mathsf{Hf}}(\ghv^{(u)}_t)$
\State user $u$ updates $\mv_t^{(u)}=\gamma \mv_{t-1}^{(u)} + \gv^{(u)}_t-\ghv^{(u)}_t $
\State user $u$ sends $\bv_t^{(u)}$ to the PS
\EndFor
\State PS decompresses all the users gradients as $\{\ghv^{(u)}_t\}_{u \in [U]}$
\State PS updates the model as $\whv_{t+1}=\whv_t+\f {\eta} U \sum_{u \in [U]} \ghv^{(u)}_t$
\EndFor
\State \Return $\whv_{T+1}$ an estimate of the optimal model $\wv^*$
\end{algorithmic}
\end{algorithm}

A summary of the parameters in the proposed approach is provided in Table. \ref{tab:algo parameters}.

\begin{table}
	\footnotesize
	\centering
	\vspace{0.04in}\caption{Recap of key parameters in alphabetical order.}
	\label{tab:algo parameters}
	\begin{tabular}{|c|c|}
		\hline
		Exponent Size & $\sf exp$ \\ \hline
		Exponent Bias & $b$ \\ \hline
	    fp Quantization & $Q_{\mathsf{fp}}(\cdot)$ \\ \hline
	    Gradient & $\gv$ \\ \hline
	    Huffman Lossless Compression & $h_{\mathsf{Hf}}(\cdot)$ \\ \hline
	    Learning Rate & $\eta$ \\ \hline
	    Mantissa Size & $\sf mant$ \\ \hline
	    Memory & $\mv$ \\ \hline
	    Memory Decay Coefficient & $\gamma$ \\ \hline
	    Sign & $\sf sgn$ \\ \hline
        Total Time & $T$ \\ \hline
	    Total Users & $U$ \\ \hline
	\end{tabular}
\end{table}

\begin{remark}{\bf Inconsequential number of users}
Note that in the approach of Algorithm \ref{alg:cap}, the number of remote users $U$  does not influence the accuracy/payload tradeoff in \eqref{eq:Lsf}. 
This is because the  algorithm parameters are not chosen as a function of $U$. In actuality, one would indeed design these hyper-parameters as a function of the number of remote users. 
 \end{remark}
Given the remark above, 
\underline{we drop the superscript ${(u)}$ in } \underline{Sec. \ref{sec:Numerical evaluations},}{ owing to the fact that the user index is inconsequential.}

\section{Numerical considerations}
\label{sec:Numerical evaluations}

In this section, we clarify various aspects of $\coiii$. 
we begin by clarifying the simulation settings, 
then revisit the three main ingredients of  $\coiii$ 
from a numerical standpoint.
Finally, we provide a plot of the overall performance of $\coiii$.

\subsection{DNN training setting}
\label{subsec:DNN train setting}
%
For our numerical evaluations, we consider the CIFAR-10 dataset classification task using the following three architectures: (i) DenseNet121, (ii) ResNet50V2, and (iii) NASNetMobile. 
For each architecture, the training is performed using SGD optimizer with a constant $\eta_{t}=0.01$ learning rate.
The rest of the configurations of the parameters and hyper-parameters used for the training are specified in Tab. \ref{tab:DNN parameters}.
%
Due to the space limitations, only the results of (iii) are shown next \footnote{The code for the gradient modeling and analysis is available at  \url{https://github.com/Chen-Zhong-Jing/CO3\_algorithm}}.

\begin{table}
	\footnotesize
	\centering
	\vspace{0.04in}\caption{Parameters and hyper-parameters used for the training of the DNN models.}
	\label{tab:DNN parameters}
	\begin{tabular}{|c|c|}
		\hline
		Dataset & CIFAR-10 \\ \hline
        Training Samples & \num{50000} \\ \hline
        Test Samples & \num{10000} \\\hline
        Optimizer & SGD \\ \hline
        Learning Rate $(\eta)$ & \num{0.01} \\ \hline
        Momentum & 0 \\ \hline
        Loss & Categorical Cross Entropy \\ \hline
        Epochs & 150 \\ \hline
        Mini-Batch Sizes & 64 \\ \hline
	\end{tabular}
\end{table}

\subsection{Gradient processing steps}
\label{sec:Gradient processing steps}

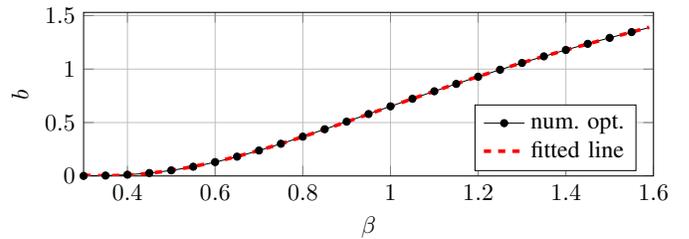
\begin{figure}
    \centering
	\begin{tikzpicture}[scale = 0.9]
    \definecolor{mycolor1}{rgb}{0.00000,0.44706,0.74118}%
    \definecolor{mycolor2}{rgb}{0.63529,0.07843,0.18431}%
    \definecolor{mycolor3}{rgb}{0.00000,0.49804,0.00000}%
    \definecolor{mycolor4}{rgb}{0.49804,0.00000,0.00000}%
    \definecolor{mycolor5}{rgb}{0.00000,0.00000,0.49804}%
    \definecolor{mycolor6}{rgb}{0.44706,0.74118,0.00000}%
    \begin{axis}[
    ymin=0,
    xmin=0.3,
    xmax=1.6,
    xlabel={$\beta$},
    ylabel={$b$},
    legend pos=south east,
    grid=both,
    height=4 cm,
    width=10cm,
    ]
        \addplot [only marks,color=black,mark=*,mark options={solid},mark repeat={5},mark size=1.5pt, smooth]
            table[x index=0, y index=1]{./Data/gennorm_beta_scalar_function.txt};
            \addlegendentry{num. opt.}
        \addplot [dashed, draw=red, line width=1.5pt, smooth]
            table[x index=0, y index=2]{./Data/gennorm_beta_scalar_function.txt};
            \addlegendentry{fitted line}
    \end{axis}
\end{tikzpicture}
    \vspace{-0.5cm}
    \caption{The relation between $\beta$ and $b$ which minimizes the $L_2$ loss for fp4 quantization}
    \label{fig:beta_scalar_function}
\end{figure}

\smallskip
\noindent
{\bf fp exponent bias:}
Let us begin by revisiting the fp exponent bias in \eqref{eq:exp bias} and argue that, when the $\gennorm$ assumption holds, then  
\ea{
b_t \approxeq  \lb 0.46 -2.85\beta + 5.37\beta^2  -2.85 \beta^3 + 0.52 \beta^4 \rb /\sigma ,
\label{eq:approximate}
}
where $\be$ is the beta parameter is the $\gennorm$ parameter corresponding to the given DNN layer, and $\sigma^2$ is the variance.
In other words, $b_t$ can be well-approximated with a polynomial that depends only on the shape parameter, once normalized by the variance.
In Fig. \ref{fig:beta_scalar_function} we plot the numerically optimized $b_t$ for fp4 as a function of $\beta$ together with the approximation in \eqref{eq:approximate} for the case in which the variance is unitary.
%
%

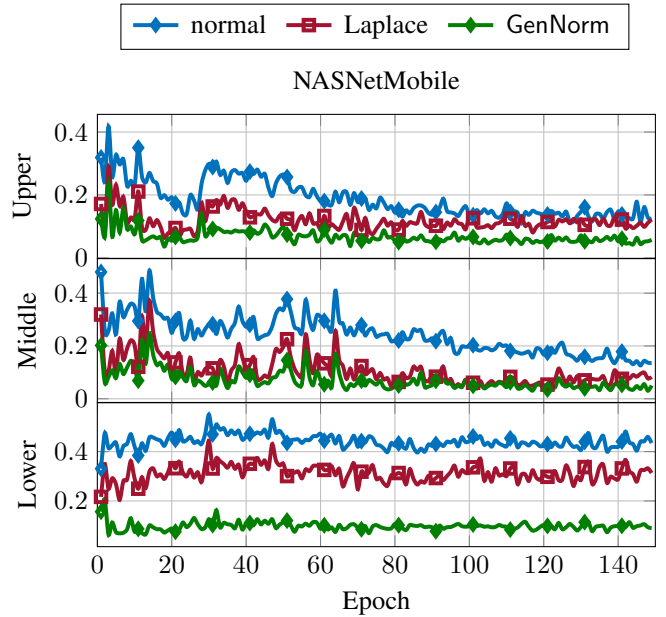
\begin{figure}
    \centering
	\begin{tikzpicture}
    \definecolor{mycolor1}{rgb}{0.00000,0.44706,0.74118}%
    \definecolor{mycolor2}{rgb}{0.63529,0.07843,0.18431}%
    \definecolor{mycolor3}{rgb}{0.00000,0.49804,0.00000}%
    \begin{groupplot}[
        group style={
            group name=my plots,
            group size=1 by 3,
            xlabels at=edge bottom,
            xticklabels at=edge bottom,
            vertical sep=0pt
        },
        height=3.5cm,
        width=9cm,
        xmax=150,
        xmin=0,
        xlabel=Epoch,
        grid=both
    ]
    \nextgroupplot[title=NASNetMobile,
    ylabel=Upper]
        \coordinate (top) at (axis cs:1,\pgfkeysvalueof{/pgfplots/ymax});
        \addplot [draw=mycolor1, line width=1.5pt, mark=diamond, mark options={solid, mycolor1}, mark repeat={10}, smooth]
            table[x index=0, y index=1]{./Data/nasnetmobile_upper_w2.txt};
            \label{plots:plot1}
        \addplot [draw=mycolor2, line width=1.5pt, mark=square, mark options={solid, mycolor2}, mark repeat={10}, smooth]
            table[x index=0, y index=2]{./Data/nasnetmobile_upper_w2.txt};
            \label{plots:plot2}
        \addplot [draw=mycolor3, line width=1.5pt, mark=diamond, mark options={solid, mycolor3}, mark repeat={10}, smooth]
            table[x index=0, y index=3]{./Data/nasnetmobile_upper_w2.txt};
            \label{plots:plot3}
    \nextgroupplot[ylabel style={align=center},
    ylabel=Middle]
        \addplot [draw=mycolor1, line width=1.5pt, mark=diamond, mark options={solid, mycolor1}, mark repeat={10}, smooth]
            table[x index=0, y index=1]{./Data/nasnetmobile_middle_w2.txt};
        \addplot [draw=mycolor2, line width=1.5pt, mark=square, mark options={solid, mycolor2}, mark repeat={10}, smooth]
            table[x index=0, y index=2]{./Data/nasnetmobile_middle_w2.txt};
        \addplot [draw=mycolor3, line width=1.5pt, mark=diamond, mark options={solid, mycolor3}, mark repeat={10}, smooth]
            table[x index=0, y index=3]{./Data/nasnetmobile_middle_w2.txt};
    \nextgroupplot[ylabel style={align=center},
    ylabel=Lower]
        \addplot [draw=mycolor1, line width=1.5pt, mark=diamond, mark options={solid, mycolor1}, mark repeat={10}, smooth]
            table[x index=0, y index=1]{./Data/nasnetmobile_lower_w2.txt};
        \addplot [draw=mycolor2, line width=1.5pt, mark=square, mark options={solid, mycolor2}, mark repeat={10}, smooth]
            table[x index=0, y index=2]{./Data/nasnetmobile_lower_w2.txt};
        \addplot [draw=mycolor3, line width=1.5pt, mark=diamond, mark options={solid, mycolor3}, mark repeat={10}, smooth]
            table[x index=0, y index=3]{./Data/nasnetmobile_lower_w2.txt};
            
    \coordinate (bot) at (axis cs:1,\pgfkeysvalueof{/pgfplots/ymin});
    \end{groupplot}
    \path (top|-current bounding box.north)--
          coordinate(legendpos)
          (bot|-current bounding box.north);
    \matrix[
        matrix of nodes,
        anchor=south,
        draw,
        inner sep=0.2em,
        draw
      ]at([yshift=1ex,xshift=23ex]legendpos)
      {
        \ref{plots:plot1}& normal &[3pt]
        \ref{plots:plot2}& Laplace &[3pt]
        \ref{plots:plot3}& $\gennorm$ &[3pt]\\};
\end{tikzpicture}
    \vspace{-0.5cm}
    \caption{$W_2$ distance between the empirical  CDF and best-fit CDF for the term   $\gv_t+\gamma \mv_{t-1}^{(u)}$ for upper, middle, and lower layers of NASNetMobile.}
    \label{fig:nasnetmobile_w2}
    \vspace{-0.5cm}
\end{figure}

\smallskip
\noindent
{\bf Huffman coding:}
After quantization, the compressed samples have a distribution corresponding to the quantized $\gennorm$ distribution. 
In this section, we wish to validate the $\gennorm$ assumption of \cite{chen2021dnn}  even when error correction is employed, that is when $\gv_t+\gamma \mv_{t-1}^{(u)}$ is considered.  
In Fig. \ref{fig:nasnetmobile_w2},  we plot the Wasserstein 2 ($W_2$) distance between the sample empirical CDF and the best-fit  CDF for three families: (i)
normal, (ii) Laplace, and (iii) generalized normal distribution. 
From Fig. \ref{fig:nasnetmobile_w2} we note that, even accounting for the fact that $\gennorm$  encompasses the normal and Laplace as special cases, the $\gennorm$ distribution offers much improved fitting of the empirical distribution of the samples to be quantized.  

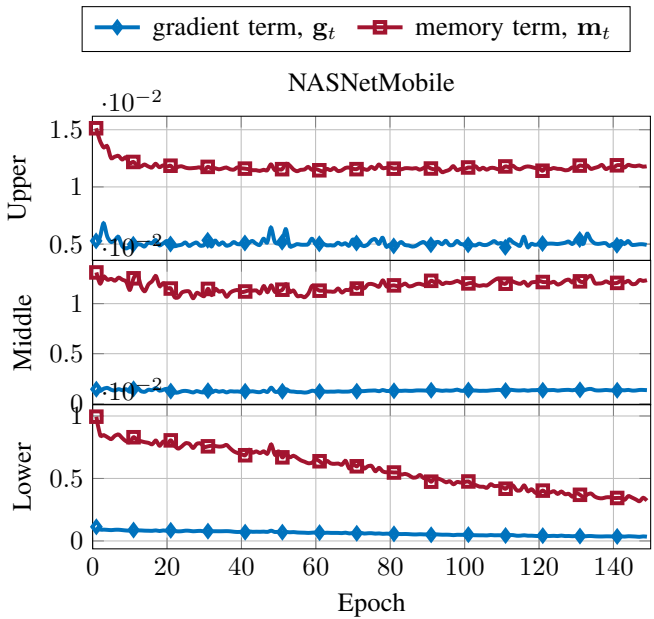
\begin{figure} 
    \centering
    \begin{tikzpicture}
    \definecolor{mycolor1}{rgb}{0.00000,0.44706,0.74118}%
    \definecolor{mycolor2}{rgb}{0.63529,0.07843,0.18431}%
    \definecolor{mycolor3}{rgb}{0.00000,0.49804,0.00000}%
    \begin{groupplot}[
        group style={
            group name=my plots,
            group size=1 by 3,
            xlabels at=edge bottom,
            xticklabels at=edge bottom,
            vertical sep=0pt
        },
        height=3.5cm,
        width=9cm,
        xmax=150,
        xmin=0,
        xlabel=Epoch,
        grid=both
    ]
    \nextgroupplot[title=NASNetMobile,
    ylabel=Upper]
        \coordinate (top) at (axis cs:1,\pgfkeysvalueof{/pgfplots/ymax});
        \addplot [draw=mycolor1, line width=1.5pt, mark=diamond, mark options={solid, mycolor1}, mark repeat={10}, smooth]
            table[x index=0, y index=1]{./Data/nasnetmobile_upper_norm.txt};
            \label{plots:plot1}
        \addplot [draw=mycolor2, line width=1.5pt, mark=square, mark options={solid, mycolor2}, mark repeat={10}, smooth]
            table[x index=0, y index=2]{./Data/nasnetmobile_upper_norm.txt};
            \label{plots:plot2}
    \nextgroupplot[ylabel style={align=center},
    ylabel=Middle]
        \coordinate (top) at (axis cs:0,\pgfkeysvalueof{/pgfplots/ymax});
         \addplot [draw=mycolor1, line width=1.5pt, mark=diamond, mark options={solid, mycolor1}, mark repeat={10}, smooth]
            table[x index=0, y index=1]{./Data/nasnetmobile_middle_norm.txt};
        \addplot [draw=mycolor2, line width=1.5pt, mark=square, mark options={solid, mycolor2}, mark repeat={10}, smooth]
            table[x index=0, y index=2]{./Data/nasnetmobile_middle_norm.txt};
    \nextgroupplot[ylabel style={align=center},
    ylabel=Lower]
        \addplot [draw=mycolor1, line width=1.5pt, mark=diamond, mark options={solid, mycolor1}, mark repeat={10}, smooth]
            table[x index=0, y index=1]{./Data/nasnetmobile_lower_norm.txt};
        \addplot [draw=mycolor2, line width=1.5pt, mark=square, mark options={solid, mycolor2}, mark repeat={10}, smooth]
            table[x index=0, y index=2]{./Data/nasnetmobile_lower_norm.txt};
            
    \coordinate (bot) at (axis cs:1,\pgfkeysvalueof{/pgfplots/ymin});
    \end{groupplot}
    \path (top|-current bounding box.north)--
          coordinate(legendpos)
          (bot|-current bounding box.north);
    \matrix[
        matrix of nodes,
        anchor=south,
        draw,
        inner sep=0.2em,
        draw
      ]at([yshift=1ex,xshift=22ex]legendpos)
      {
        \ref{plots:plot1}& gradient term, $\gv_t$ &[3pt]
        \ref{plots:plot2}& memory  term, $\mv_t$ &[3pt]\\};
\end{tikzpicture}
    \vspace{-0.5cm}
    \caption{$L_1$ norm of gradient and error term for upper, middle, and lower layers from the NASNetMobile when $\gamma=0.9$.}
    \label{fig:nasnetmobile_l1norm}
    \vspace{-0.5cm}
\end{figure}

\smallskip
\noindent
{\bf Error Correction:}
%
%
Next, in Fig. \ref{fig:nasnetmobile_l1norm}, we investigate the magnitude of the memory term, $\mv_t$, versus the gradient term, $\gv_t$, in the error feedback mechanisms with $\gamma=0.9$. 
%
%
We note that the relative amplitude of $\mv_t$ and $\gv_t$ is rather stable for this choice of $\gamma$, with exceptions of the lower layers,  where the memory is vanishing. 
Finally, we choose the $L_1$ norm in Fig. \ref{fig:nasnetmobile_l1norm} following
\cite{karimireddy2019error}.

\subsection{Overall performance}
\label{sec:Overall performance}
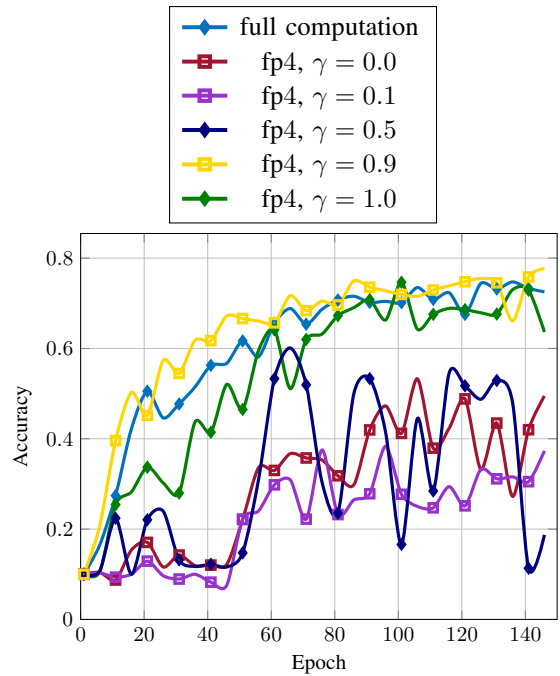
\begin{figure}
    \centering
	\begin{tikzpicture}[scale = 0.8]
    \definecolor{mycolor1}{rgb}{0.00000,0.44706,0.74118}%
    \definecolor{mycolor2}{rgb}{0.63529,0.07843,0.18431}%
    \definecolor{mycolor3}{rgb}{0.00000,0.49804,0.00000}%
    \definecolor{mycolor4}{rgb}{0.60000,0.19608,0.80000}%
    \definecolor{mycolor5}{rgb}{0.00000,0.00000,0.50196}%
    \definecolor{mycolor6}{rgb}{1.00000,0.84314,0.00000}%
    \begin{axis}[
    ymin=0,
    xmin=0,
    xmax=150,
    xlabel={Epoch},
    ylabel={Accuracy},
    grid=both,
    height=8cm,
    width=9.5cm,
    ]
    \coordinate (top) at (axis cs:1,\pgfkeysvalueof{/pgfplots/ymax});
        \addplot [draw=mycolor1, line width=1.5pt, mark=diamond, mark options={solid, mycolor1}, mark repeat={2}, smooth, each nth point=5, filter discard warning=false, unbounded coords=discard ]
            table[x index=0, y index=1]{./Data/nasnetmobile_ef.txt};
            \label{plots:plot1}
        \addplot [draw=mycolor2, line width=1.5pt, mark=square, mark options={solid, mycolor2}, mark repeat={10}, smooth, mark repeat={2}, smooth, each nth point=5, filter discard warning=false, unbounded coords=discard ]
            table[x index=0, y index=2]{./Data/nasnetmobile_ef.txt};
            \label{plots:plot2}
        \addplot [draw=mycolor3, line width=1.5pt, mark=diamond, mark options={solid, mycolor3}, mark repeat={10}, smooth, mark repeat={2}, smooth, each nth point=5, filter discard warning=false, unbounded coords=discard ]
            table[x index=0, y index=3]{./Data/nasnetmobile_ef.txt};
            \label{plots:plot3}
        \addplot [draw=mycolor4, line width=1.5pt, mark=square, mark options={solid, mycolor4}, mark repeat={10}, smooth, mark repeat={2}, smooth, each nth point=5, filter discard warning=false, unbounded coords=discard ]
            table[x index=0, y index=4]{./Data/nasnetmobile_ef.txt};
            \label{plots:plot4}
        \addplot [draw=mycolor5, line width=1.5pt, mark=diamond, mark options={solid, mycolor5}, mark repeat={10}, smooth, mark repeat={2}, smooth, each nth point=5, filter discard warning=false, unbounded coords=discard ]
            table[x index=0, y index=5]{./Data/nasnetmobile_ef.txt};
            \label{plots:plot5}
        \addplot [draw=mycolor6, line width=1.5pt, mark=square, mark options={solid, mycolor6}, mark repeat={10}, smooth, mark repeat={2}, smooth, each nth point=5, filter discard warning=false, unbounded coords=discard]
            table[x index=0, y index=6]{./Data/nasnetmobile_ef.txt};
            \label{plots:plot6}
    \coordinate (bot) at (axis cs:1,\pgfkeysvalueof{/pgfplots/ymin});
    \end{axis}
    \path (top|-current bounding box.north)--
          coordinate(legendpos)
          (bot|-current bounding box.north);
    \matrix[
        matrix of nodes,
        anchor=south,
        draw,
        inner sep=0.2em,
        draw
      ]at([yshift=1ex,xshift=21ex]legendpos)
      {
        \ref{plots:plot1}& full computation &[3pt]\\
        \ref{plots:plot2}& fp$4$, $\gamma=0.0$ &[3pt]\\
        \ref{plots:plot4}& fp$4$, $\gamma=0.1$ &[3pt]\\
        \ref{plots:plot5}& fp$4$, $\gamma=0.5$ &[3pt]\\
        \ref{plots:plot6}& fp$4$, $\gamma=0.9$ &[3pt]\\
        \ref{plots:plot3}& fp$4$, $\gamma=1.0$ &[3pt]\\};
\end{tikzpicture}
    \vspace{-0.2cm}
    \caption{Test accuracy of NASNetMobile. The communication overhead is $\Rsf_{fp4} = \num{2.15e12}$ bits.}
    \label{fig:accuracy}
    \vspace{-0.5cm}
\end{figure}

We conclude the paper with a plot of the overall performance of the $\coiii$: see Fig. \ref{fig:accuracy}.
 In Fig. \ref{fig:accuracy}, we plot the performance for $[\sf sgn \ \sf mant \ \sf exp]=[1 \ 2 \ 1]$, 
 %
 %
and various values of $\gamma$.
Through our experimentation, we notice that with $\gamma=0.9$, the proposed $\coiii$ can provide performance comparable to the full SGN computation while requiring a significantly less communication resource.


\section{Conclusion}
In this paper, we propose $\coiii$, a novel algorithm for communication-efficient distributed DNN training, which is comprised of three fundamental gradient processing steps: (i) floating point conversion, (ii) lossless compression, and (iii) error correction.
Extensive simulations have been provided to demonstrate that $\coiii$ has excellent performance at a very reasonable communication payload. 

\bibliographystyle{IEEEtran}
\bibliography{IEEEabrv,ICC_zhong_jin }


\end{document}